\documentclass[a4paper,11pt]{article}
\usepackage{graphicx}
\usepackage{amsmath}
\usepackage{amssymb}
\usepackage{hyperref}
\usepackage{booktabs}
\usepackage{caption}
\usepackage{geometry}
\usepackage[T1]{fontenc}
\usepackage{subcaption}

\usepackage{graphicx}
\usepackage{bbm}
\usepackage{algorithmic}
\usepackage{xcolor}
\usepackage{algorithm}
\usepackage{multirow}
\usepackage{longtable}
\usepackage{graphicx}

\usepackage{tikz}
\usepackage{pgfplots}
\usepackage{multirow}
\usepackage{amssymb}
\tikzset{
    axis break gap/.initial=0mm
}
\usepackage{booktabs}
\usepackage{array}
\usepackage{float} 
\usepackage{subcaption} 
\usepackage{authblk}
\usepackage{amsthm}
\usepackage{booktabs}      
\usepackage{xcolor}        
\usepackage{colortbl}      
\usepackage{graphicx}      
\usepackage{multirow}      
\usepackage{rotating}

\usepackage{multirow}
\usepackage{booktabs}
\usepackage{adjustbox}
\theoremstyle{definition}
\newtheorem{definition}{Definition}[section]  

\theoremstyle{remark}

\usepackage{natbib}
\setcitestyle{authoryear,open={(},close={)}}

\title{AxelSMOTE: An Agent-Based Oversampling Algorithm for Imbalanced Classification}

\author[1]{
Sukumar Kishanthan
}

\author[2]{
Asela Hevapathige\thanks{Corresponding author}
}

\affil[1]{Dialog Axiata PLC, Colombo, Sri Lanka, Email: kishanthansukumar@gmail.com}

\affil[2]{School of Computing, Australian National University, Canberra, Australia, Email: asela.hevapathige@anu.edu.au}

\date{}  

\begin{document}
\maketitle

\section*{Abstract}

Class imbalance in machine learning poses a significant challenge, as skewed datasets often hinder performance on minority classes. Traditional oversampling techniques, which are commonly used to alleviate class imbalance, have several drawbacks: they treat features independently, lack similarity-based controls, limit sample diversity, and fail to manage synthetic variety effectively. To overcome these issues, we introduce AxelSMOTE, an innovative agent-based approach that views data instances as autonomous agents engaging in complex interactions. Based on Axelrod's cultural dissemination model, AxelSMOTE implements four key innovations: (1) trait-based feature grouping to preserve correlations; (2) a similarity-based probabilistic exchange mechanism for meaningful interactions; (3) Beta distribution blending for realistic interpolation; and (4) controlled diversity injection to avoid overfitting. Experiments on eight imbalanced datasets demonstrate that AxelSMOTE outperforms state-of-the-art sampling methods while maintaining computational efficiency.

\textbf{Keywords:} class imbalance, agent-based interactions, oversampling, statistical physics, algorithms

\section{Introduction}

Class imbalance refers to the phenomenon where the distribution of classes in a dataset is biased towards specific classes \citep{leevy2018survey,aguiar2024survey}. It is identified as a challenging problem in data mining and machine learning, as machine learning and deep learning classifiers tend to perform suboptimally under imbalanced data, specifically towards minority class instances \citep{altalhan2025imbalanced,ghosh2024class}. This problem cannot be ignored as many real-world and engineering datasets naturally have imbalanced class distributions due to their real-world characteristics \citep{kishanthan2025deep,chen2024survey}.

Researchers have explored various mechanisms to alleviate class imbalance issues for decades, leading to the development of diverse techniques to improve classifier performance on minority classes. These techniques include data resampling, where sampling is performed to balance class distributions \citep{carvalho2025resampling,zhao2024resampling}, algorithm-level approaches that modify the learning model to be more sensitive to minority classes \citep{araf2024cost,farhadpour2024selecting}, and hybrid approaches that combine both techniques \citep{shin2024towards,goswami2024literature}. Among these methods, data resampling is commonly preferred due to its model-agnostic nature, simplicity, and effectiveness in practice. Specifically, data oversampling, which increases the number of minority samples, has been empirically shown to work better for real-world datasets.  Unlike undersampling, which reduces the total amount of training data by discarding instances from the majority class, oversampling increases the dataset size by maintaining all original instances and adding more examples. This approach allows the model to learn more accurate decision boundaries with enhanced generalizability. Additionally, compared to hybrid sampling methods, oversampling is simpler and easier to interpret, as it avoids the complexities and additional hyperparameters associated with balancing multiple sampling techniques. 

Commonly used oversampling techniques include diverse methods such as SMOTE \cite{chawla2002smote}, SVMSMOTE \cite{tang2008svms}, BorderlineSMOTE \cite{han2005borderline}, and ADASYN \cite{he2008adasyn}, which focus on generating synthetic minority data instances by fusing the existing minority data. Nonetheless, these existing oversampling methods suffer from several limitations. 

\begin{itemize}
\item \textbf{Feature Independence Assumption:} Traditional oversampling methods treat features independently during synthetic sample generation, potentially breaking important feature correlations and semantic relationships that exist within the original data.

\item \textbf{Lack of Similarity-Based Generation Control:} Most existing oversampling approaches employ simple interpolation between samples without adequately considering whether the selected samples are sufficiently similar for meaningful synthetic generation, which can result in unrealistic synthetic instances.

\item \textbf{Deterministic Generation Processes:} The generation processes in these methods tend to be relatively deterministic, following predictable patterns that may limit the diversity of the generated samples and reduce the robustness of the approach.

\item \textbf{Limited Diversity Control Mechanisms:} Traditional approaches generally lack sophisticated mechanisms for controlling diversity in synthetic samples, which can lead to overfitting to the training data and reduced generalization capability of the resulting models.
\end{itemize}

In this work, we aim to address these limitations.  We propose a novel oversampling approach that reconceptualises data instances as autonomous agents capable of complex interactions during synthetic sample generation. We observe that modelling data instances as agents enables the capture of sophisticated relationships and dependencies that traditional methods overlook. Specifically, we adopt Axelrod's cultural dissemination model \citep{axelrod1997dissemination} as the theoretical foundation for our method, as it offers several key advantages: it naturally handles multi-dimensional feature interactions through the concept of cultural traits, incorporates similarity-based exchange mechanisms that ensure meaningful interactions between compatible instances, introduces probabilistic elements that add controlled randomness to the generation process, and provides explicit parameters for managing diversity and interaction patterns. Axelrod's model is particularly well-suited for oversampling because it was originally designed to explain how similar entities influence each other while maintaining diversity, which directly parallels the challenge of generating realistic synthetic samples that are similar to existing minority instances while avoiding exact replication. By leveraging this agent-based cultural exchange paradigm, our approach systematically addresses each of the identified gaps while providing a theoretically grounded and interpretable framework for synthetic sample generation. We call our oversampling method AxelSMOTE (\underline{Axel}ord \underline{S}ynthetic \underline{M}inority \underline{O}versampling \underline{TE}chnique). The main contributions of our work are as follows:

\begin{itemize}
\item \textbf{A Novel Agent-Based Perspective on Oversampling}: We propose reconceptualizing data instances as autonomous agents capable of complex interactions, introducing Axelrod's dissemination model as a theoretical foundation for synthetic sample generation in imbalanced datasets.

\item \textbf{Trait-Based Feature Grouping for Correlation Preservation}: We develop the concept of feature traits that partition features into semantically related groups, ensuring that correlated features are modified collectively during synthetic generation, thereby preserving important intra-trait relationships that traditional methods often break.

\item \textbf{Similarity-Based Probabilistic Exchange Mechanism}: We introduce a control mechanism that combines similarity thresholds with probabilistic influence rates to ensure meaningful trait exchange occurs only between sufficiently compatible instances, preventing unrealistic synthetic sample generation.

\item \textbf{Empirical Performance}: Extensive experiments demonstrate that AxelSMOTE consistently outperforms state-of-the-art sampling methods across multiple imbalanced datasets.
\end{itemize}

\section{Related Work}

There has been a plethora of works targeting alleviating class imbalance. These works can be divided into three categories: resampling methods, algorithmic methods, and hybrid approaches.  Data resampling techniques aim to decrease the imbalance in class distribution using various sampling methods \citep{gurcan2024learning}. On the other hand, algorithmic methods focus on enhancing the classifier through modifying the loss function or model architecture to increase its sensitivity to minority classes, where hybrid methods combine both data reassembly and algorithmic methods \citep{krawczyk2016learning}.

\paragraph{Data Resampling:} Data resampling can be done either through minority oversampling, majority undersampling, or a combination of both these approaches. Minority oversampling increases the number of instances in the minority class, while majority undersampling reduces majority data instances to match the distribution of the minority class. Hybrid sampling strategically combines both minority oversampling and majority undersampling. Our research focuses on data resampling techniques by proposing an oversampling method to address class imbalance. 

\begin{table}[htbp]
\centering
\caption{Comparison of AxelSMOTE with existing oversampling methods}
\label{tab:method_comparison}
\resizebox{\textwidth}{!}{%
\begin{tabular}{l|c|c|c|c|c|c|c}
\toprule
\textbf{Method} & 
\textbf{Feature} & 
\textbf{Similarity-based} & 
\textbf{Probabilistic} & 
\textbf{Diversity} & 
\textbf{Theoretical} & 
\textbf{Computational} & 
\textbf{Stability} \\
& 
\textbf{Correlation} & 
\textbf{Generation} & 
\textbf{Generation} & 
\textbf{Control} & 
\textbf{Foundation} & 
\textbf{Efficiency} & 
\\
& 
\textbf{Preservation} & 
\textbf{Control} & 
& 
& 
& 
& 
\\
\midrule
\multicolumn{8}{c}{\textit{Traditional Methods}} \\
\midrule
SMOTE & None & None & None & None & None & High & High \\
SVMSMOTE & None & Moderate & None & None & Low & High & High \\
BorderlineSMOTE & None & Moderate & None & None & Low & High & High \\
ADASYN & None & Moderate & None & Low & Low & High & High \\
K-means SMOTE & None & Moderate & None & Low & Low & High & High \\
\midrule
\multicolumn{8}{c}{\textit{Deep Learning Methods}} \\
\midrule
GAN-based & Low & None & High & High & Moderate & Low & Low \\
VAE-based & Low & None & High & Moderate & Moderate & Low & Moderate \\
DeepSMOTE & Low & None & Moderate & Moderate & Moderate & Low & Moderate \\
\midrule
\multicolumn{8}{c}{\textit{Agent-based Method}} \\
\midrule
\textbf{AxelSMOTE} & \textbf{High} & \textbf{High} & \textbf{High} & \textbf{High} & \textbf{High} & \textbf{High} & \textbf{High} \\
\bottomrule
\end{tabular}
}
\end{table}

\paragraph{Minority Oversampling:} One of the early works that explored minority oversampling was SMOTE, where the synthetic samples were generated through linear interpolation of minority instances \citep{chawla2002smote}. Since then, several sophisticated variants have been proposed to address SMOTE's limitations. SVMSMOTE\citep{tang2008svms} employs support vector machine concepts to position synthetic samples strategically along class boundaries to have better classifier discrimination capabilities. Borderline SMOTE \citep{han2005borderline} identifies that the minority instances located at classification margins are more prone to be misclassified, and prioritises them in the oversampling process.  K-means SMOTE \citep{douzas2018improving} introduces a clustering-based enhancement where minority data instances are first clustered before being used to generate synthetic data within each cluster. This approach helps to better preserve the original data structure and reduces the likelihood of producing noisy or unrealistic synthetic samples. ADASYN \citep{he2008adasyn} is an adaptive synthetic data generation approach that focuses on regions where the minority class is more difficult to learn. It does this by generating synthetic samples for each minority instance based on its difficulty level.  Feature type extensions are proposed to SMOTE through SMOTE-N and SMOTE-NC, which provide tailored handling mechanisms for categorical and continuous feature datasets \citep{chawla2002smote}. Further, there have been deep learning-based oversampling methods with have learnable parameters that adaptively generate synthetic minority class instances by adapting to the underlying data distribution \citep{ando2017deep,dablain2022deepsmote,mullick2019generative,engelmann2021conditional,karunasingha2023oc,kishanthan2025deep}. However, such methods are mostly based upon Generative Adversarial Networks (GANs), and Variational Autoencoders (VAEs), which is prone to high computational costs and issues like model collapse \citep{zhang2021mode,barsha2025depth}.

Notably, existing oversampling methods do not account for feature correlations and diversity control in the generated synthetic data, thereby limiting the realism and generalizability of the generated data. A high-level comparison of AxelSMOTE with existing oversampling methods is provided in Table \ref{tab:method_comparison}.

\section{Background}

In this section, we introduce the 
Axelrod's model of cultural dissemination \citep{axelrod1997dissemination}. It is a social dynamics model in statistical physics that studies how cultural diversity persists with social influence. The model is based on two key principles: (1) people are more likely to interact with others who share similar cultural traits, and (2) these interactions tend to increase their cultural similarity, creating a self-reinforcing cycle. 

\begin{algorithm}
\caption{Axelrod Cultural Dissemination Model}
\label{alg:axelrod_original}
\begin{algorithmic}[1]
\REQUIRE Grid size $L$, number of features $f$, number of traits per feature $q$
\ENSURE Final cultural configuration $\mathcal{G}$

\STATE Initialize $L \times L$ grid $\mathcal{G}$ of agents
\FOR{each agent $i$ in $\mathcal{G}$}
    \FOR{each feature $j = 1$ to $f$}
        \STATE $x_i^{(j)} \leftarrow \text{Random}(0, q-1)$ \COMMENT{Assign random trait}
    \ENDFOR
\ENDFOR

\REPEAT
    \STATE $\text{changed} \leftarrow \text{false}$
    \STATE Select random active agent $i$ from $\mathcal{G}$
    \STATE Select random neighbor $k$ from adjacent cells of $i$
    
    \STATE Calculate cultural similarity: $\text{similarity} = \frac{n_{ik}}{f}$
    \STATE where $n_{ik} = \sum_{j=1}^{f} \mathbbm{1}[x_i^{(j)} = x_k^{(j)}]$ \COMMENT{Number of shared traits}
    
    \STATE Generate random number $p \sim U(0,1)$
    \IF{$p < \text{similarity}$}
        \STATE Find differing features: $\mathcal{D} = \{j : x_i^{(j)} \neq x_k^{(j)}\}$
        \IF{$|\mathcal{D}| > 0$}
            \STATE Randomly select feature $j^* \in \mathcal{D}$
            \STATE $x_i^{(j^*)} \leftarrow x_k^{(j^*)}$ \COMMENT{Agent $i$ adopts trait from agent $k$}
            \STATE $\text{changed} \leftarrow \text{true}$
        \ENDIF
    \ENDIF
\UNTIL{$\text{changed} = \text{false}$ for sufficient iterations}

\RETURN Final cultural configuration $\mathcal{G}$
\end{algorithmic}
\label{algo:axelord_model}
\end{algorithm}

Axelrod's cultural dissemination model operates on a spatial lattice $\mathcal{G}$ of size $L \times L$, where each cell represents a stationary agent. Every agent $i$ possesses a culture characterized by a vector $\mathbf{x}_i = (x_i^{(1)}, x_i^{(2)}, \ldots, x_i^{(f)})$ of $f$ cultural features, where each feature $x_i^{(j)} \in \{0, 1, \ldots, q-1\}$ can take one of $q$ possible traits. The cultural similarity between two agents $i$ and $k$ is defined as $\text{sim}(i,k) = \frac{n_{ik}}{f}$, where $n_{ik} = \sum_{j=1}^{f} \mathbbm{1}[x_i^{(j)} = x_k^{(j)}]$ represents the number of features for which both agents share identical traits. Agents interact only with their spatial neighbors (i.e., adjacent cells in the lattice), and the probability of interaction is directly proportional to their cultural similarity. During interaction, an active agent selects one feature where it differs from a passive neighbor and adopts the neighbor's trait for that feature, thereby increasing their cultural similarity and making future interactions more likely. The algorithm of Axelord cultural dissemination model is depicted in Algorithm \ref{algo:axelord_model}.

\section{Methodology}

\begin{figure}[htbp]
    \centering
    \includegraphics[width=1.0\textwidth]{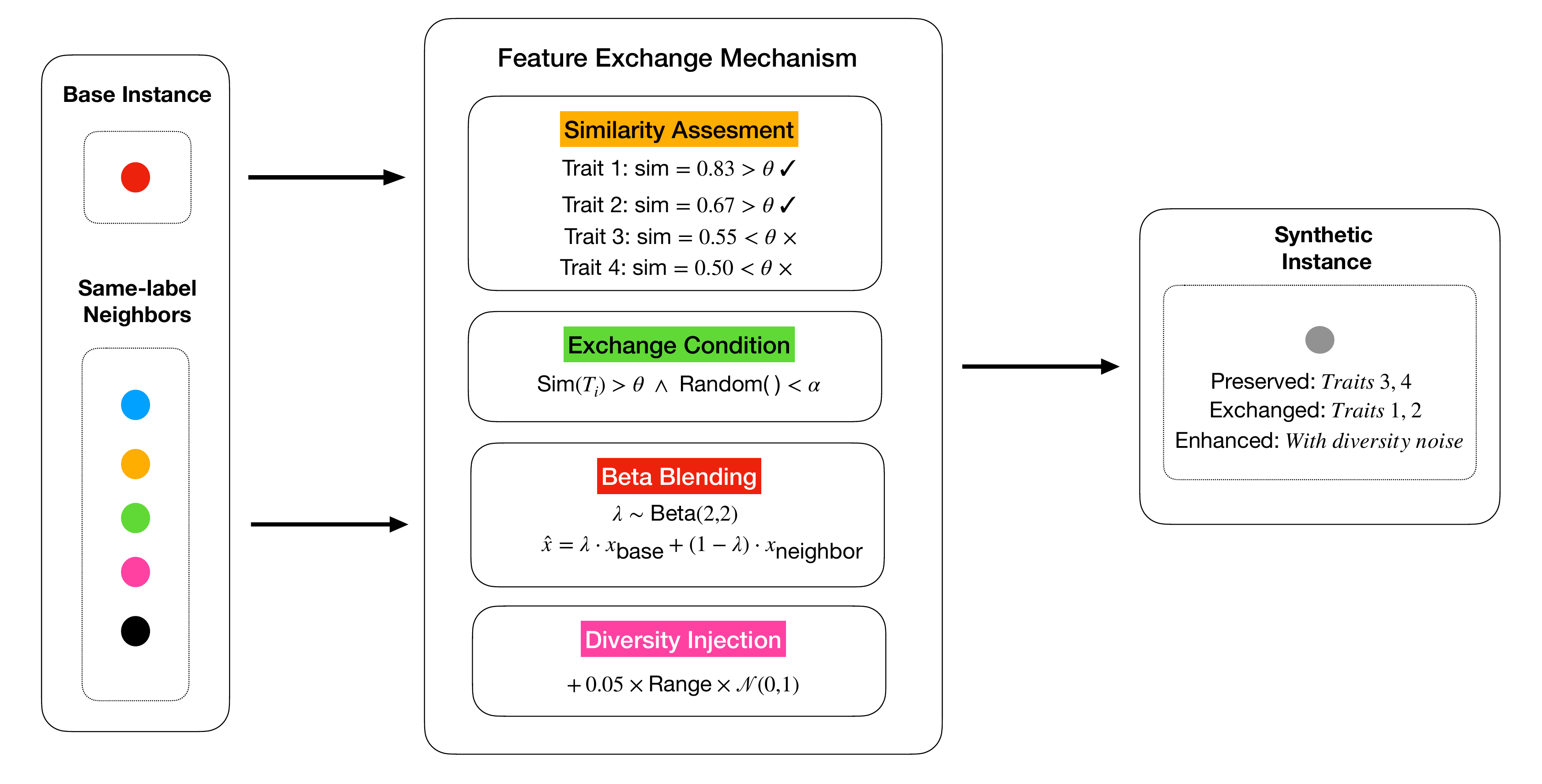}
    \caption{High-level Overview of AxelSMOTE synthetic sample generation process. Our method treats data instances as autonomous agents with traits. It starts with a base agent and its k-nearest neighbours, performing similarity-based feature exchange where traits are exchanged probabilistically based on thresholds and influence rates with Beta distribution blending. This process generates synthetic samples that maintain intra-trait correlations while adding controlled diversity through Gaussian noise. }
    \label{fig:highlevel_architecture}
\end{figure}

\subsection{Preliminaries}

Consider a labeled training dataset $\mathcal{D} = \{(\mathbf{x}_i, l_i)\}_{i=1}^n$, where $\mathbf{x}_i \in \mathbb{R}^d$ denotes the $i$-th sample vector with $d$ dimensions, and $l_i \in \mathcal{L} = \{1, 2, \ldots, m\}$ represents its corresponding class label across $m$ distinct classes.

For a given sample $\mathbf{x}_i$ and a positive integer $k$, we define the $k$-nearest neighbor set as $\mathcal{N}_k(\mathbf{x}_i) = \{\mathbf{x}_{j_1}, \mathbf{x}_{j_2}, \ldots, \mathbf{x}_{j_k}\}$, where each neighbor $\mathbf{x}_{j_\ell} \in \mathcal{D}$ satisfies $l_{j_\ell} = l_i$ and the ordering constraint $\|\mathbf{x}_i - \mathbf{x}_{j_1}\|_2 \leq \|\mathbf{x}_i - \mathbf{x}_{j_2}\|_2 \leq \cdots \leq \|\mathbf{x}_i - \mathbf{x}_{j_k}\|_2 \leq \|\mathbf{x}_i - \mathbf{x}_{j'}\|_2$ for any other same-class sample $\mathbf{x}_{j'} \notin \mathcal{N}_k(\mathbf{x}_i)$.

\subsection{Class Imbalance Problem Formulation}

A classification task exhibits class imbalance when the distribution of samples across classes is significantly skewed. Formally, let $\mathcal{S}_c = \{\mathbf{x}_i \in \mathcal{D} : l_i = c\}$ represent the subset of samples belonging to class $c$. The dataset is considered imbalanced if:

\begin{equation}
\exists c^* \in \mathcal{L} : |\mathcal{S}_{c^*}| < \gamma \cdot \max_{c \in \mathcal{L}} |\mathcal{S}_c|
\end{equation}

where $\gamma \in (0, 1)$ is a threshold parameter (typically $\gamma = 0.5$ or smaller). Classes satisfying this condition are termed \emph{minority classes}, while the class with maximum cardinality constitutes the \emph{majority class}.

The degree of imbalance is commonly measured using the imbalance ratio:
\begin{equation}
\text{IR} = \frac{\max_{c \in \mathcal{L}} |\mathcal{S}_c|}{\min_{c \in \mathcal{L}} |\mathcal{S}_c|}
\end{equation}

The fundamental challenge in imbalanced classification is to learn a mapping function $h: \mathbb{R}^d \rightarrow \mathcal{L}$ that maintains high predictive performance across all classes, particularly the underrepresented minority classes.

\subsection{Synthetic Oversampling Framework}

Oversampling techniques address class imbalance by augmenting minority classes with artificially generated samples. We formalize this process through a synthetic generation operator $\Phi: \mathcal{D} \rightarrow \mathcal{D}^+$, where $\mathcal{D}^+$ represents the expanded dataset:

\begin{equation}
\mathcal{D}^+ = \mathcal{D} \cup \bigcup_{c \in \mathcal{M}} \Phi_c(\mathcal{S}_c)
\end{equation}

Here, $\mathcal{M} \subseteq \mathcal{L}$ denotes the set of minority classes requiring augmentation, and $\Phi_c(\mathcal{S}_c)$ generates synthetic samples for class $c$ based on its existing instances.

At the instance level, synthetic sample generation can be expressed as a neighborhood-based transformation:
\begin{equation}
\hat{\mathbf{x}} = \Psi(\mathbf{x}_{\text{seed}}, \mathcal{N}_k(\mathbf{x}_{\text{seed}}))
\end{equation}

where $\mathbf{x}_{\text{seed}} \in \mathcal{S}_c$ serves as a reference point, and $\Psi(\cdot)$ represents a synthesis function that combines information from the seed sample and its neighborhood to produce a new synthetic instance $\hat{\mathbf{x}}$.

\subsection{Proposed Oversampling Approach}

Inspired by Axelrod's model of cultural dissemination, we propose a novel oversampling approach AxelSMOTE that models synthetic sample generation as a cultural exchange process controlled by four key hyperparameters: \textbf{k} representing the number of nearest neighbors to consider, \textbf{t} denoting the number of feature traits per group, \textbf{$\theta$} as the similarity threshold for cultural exchange, and \textbf{$\alpha$} representing the influence rate (probability of exchange when similarity condition is met). The approach incorporates the following principles:
\begin{itemize}
\item Feature Coherence: Features are grouped into traits that are exchanged together
\item Similarity-Based Interaction: Feature exchange occurs based on similarity thresholds
\item Probabilistic Influence: Exchange probability depends on feature compatibility
\end{itemize}

\subsubsection{Concept Definitions}

We start with the following definitions related to the proposed approach.

\begin{definition}[Feature Traits]
We partition the $d$ features in each data vector into $t$ feature traits, where each trait $T_j$ represents a subset of features that are conceptually related and should be modified together:
\begin{equation}
T_j = \left\{d_{(j-1) \cdot \lfloor d/t \rfloor + 1}, \ldots, d_{\min(j \cdot \lfloor d/t \rfloor, d)}\right\}, \quad j = 1, 2, \ldots, t
\end{equation}
\end{definition}

\begin{definition}[Feature Similarity]
For feature trait $T_j$ between two data instances $\mathbf{x}_i$ and $\mathbf{x}_\ell$, the feature similarity is defined as:
\begin{equation}
\text{sim}_j(\mathbf{x}_i, \mathbf{x}_\ell) = 1 - \frac{1}{|T_j|} \sum_{q \in T_j} |\mathbf{x}_i^{(q)} - \mathbf{x}_\ell^{(q)}|
\end{equation}
where $\mathbf{x}^{(q)}$ denotes the $q$-th feature of sample $\mathbf{x}$.
\end{definition}

\subsubsection{Synthetic Sample Generation Process}

To generate a synthetic sample $\hat{\mathbf{x}}$, we randomly select a base sample $\mathbf{x}_b \in \mathcal{S}_c$ and identify its $k$ nearest same-class neighbors. The synthetic sample is initialized as a copy of the base sample: 
\begin{equation}
\hat{\mathbf{x}} \leftarrow \mathbf{x}_b
\end{equation}

The core innovation of our approach lies in the feature exchange process, where features are grouped into feature traits and exchanged probabilistically based on similarity. For each feature trait $T_j$, we randomly select up to $k$ neighbors from $\mathcal{N}_k(\mathbf{x}_b)$:
\begin{equation}
\mathcal{N}'_j \subseteq \mathcal{N}_k(\mathbf{x}_b) 
\end{equation}

 Then, for each selected neighbor $\mathbf{x}_n \in \mathcal{N}'_j$, cultural exchange occurs if two conditions are met:

\begin{itemize}
\item Similarity Condition: The feature similarity exceeds the threshold:
\begin{equation}
\text{sim}_j(\mathbf{x}_b, \mathbf{x}_n) > \theta
\end{equation}

\item Probabilistic Condition: A random event occurs with probability $\alpha$:
\begin{equation}
U(0,1) < \alpha
\end{equation}
\end{itemize}

When both conditions are satisfied, trait $T_j$ undergoes feature exchange. When feature exchange occurs for trait $T_j$, we perform realistic blending between the base sample and the neighbor. We sample a blending ratio from a Beta distribution with fixed parameters for balanced mixing:
\begin{equation}
\lambda \sim \text{Beta}(2, 2)
\end{equation}

The Beta(2,2) distribution creates a symmetric, bell-shaped distribution over $[0,1]$ that favors moderate blending over extreme values. The trait features are then updated as:
\begin{equation}
\hat{\mathbf{x}}^{(p)} \leftarrow \lambda \cdot \mathbf{x}_b^{(p)} + (1-\lambda) \cdot \mathbf{x}_n^{(p)}, \quad \forall p \in T_j
\end{equation}

where $p$ represents a feature index within a specific trait. This ensures that all features within a feature trait are modified collectively, preserving intra-trait correlations. To further enhance the diversity of synthetic samples and prevent overfitting to the training data, we inject controlled variation with probability $\alpha$ (the same influence rate parameter). For each trait $T_j$ that underwent feature exchange, we apply small-scale Gaussian noise:

\begin{equation}
\hat{\mathbf{x}}^{(p)} \leftarrow \hat{\mathbf{x}}^{(p)} + 0.05 \cdot R_p \cdot \epsilon_p
\end{equation}

where $R_p = \max_{\mathbf{x} \in \mathcal{S}_c} \mathbf{x}^{(p)} - \min_{\mathbf{x} \in \mathcal{S}_c} \mathbf{x}^{(p)}$ is the feature range and $\epsilon_p \sim \mathcal{N}(0, 1)$. The pseudocode of the proposed approach is depicted in Algorithm \ref{alg:sampling_algorithm}. The high-level architecture of AxelSMOTE is depicted in Figure \ref{fig:highlevel_architecture}.

\begin{algorithm}
\caption{AxelSMOTE}
\label{alg:axelrod_simple}
\begin{algorithmic}[1]
\REQUIRE Dataset $\mathcal{D} = \{(\mathbf{x}_i, l_i)\}_{i=1}^n$, hyperparameters $(k, t, \theta, \alpha)$
\ENSURE Augmented dataset $\mathcal{D}^+$

\STATE Initialize $\mathcal{D}^+ \leftarrow \mathcal{D}$
\STATE Compute class distributions and identify minority classes
\STATE Partition features into $t$ feature traits: $\{T_1, T_2, \ldots, T_t\}$

\FOR{each minority class $c$ requiring oversampling}
    \STATE Extract class samples: $\mathcal{S}_c = \{\mathbf{x}_i : l_i = c\}$
    \IF{$|\mathcal{S}_c| \leq 1$}
        \STATE \textbf{continue} \COMMENT{Skip insufficient samples}
    \ENDIF
    \STATE Determine number of synthetic samples to generate: $s_c$
    
    \FOR{$j = 1$ to $s_c$}
        \STATE Randomly select base sample: $\mathbf{x}_b \leftarrow \text{Random}(\mathcal{S}_c)$
        \STATE Find same-label neighbors of $\mathbf{x}_b$: $\mathcal{N}_k(\mathbf{x}_b)$
        \STATE Initialize synthetic sample: $\hat{\mathbf{x}} \leftarrow \mathbf{x}_b$
        \STATE Initialize exchange tracker: $\mathcal{E} \leftarrow \emptyset$
        
        \FOR{each feature trait $T_i$, $i = 1$ to $t$}
            \STATE Randomly select up to $k$ neighbors: $\mathcal{N}'_i \leftarrow \text{RandomSubset}(\mathcal{N}_k(\mathbf{x}_b), k)$
            
            \FOR{each neighbor $\mathbf{x}_n \in \mathcal{N}'_i$}
                \STATE Compute similarity: $\text{sim}_i(\mathbf{x}_b, \mathbf{x}_n) = 1 - \frac{1}{|T_i|} \sum_{p \in T_i} |\mathbf{x}_b^{(p)} - \mathbf{x}_n^{(p)}|$
                
                \IF{$\text{sim}_i(\mathbf{x}_b, \mathbf{x}_n) > \theta$ and $\text{Random}(0,1) < \alpha$}
                    \STATE Sample blend ratio: $\lambda \sim \text{Beta}(2, 2)$
                    \FOR{each feature $p \in T_i$}
                        \STATE $\hat{\mathbf{x}}^{(p)} \leftarrow \lambda \cdot \mathbf{x}_b^{(p)} + (1-\lambda) \cdot \mathbf{x}_n^{(p)}$
                    \ENDFOR
                    \STATE $\mathcal{E} \leftarrow \mathcal{E} \cup \{T_i\}$ \COMMENT{Mark trait as exchanged}
                \ENDIF
            \ENDFOR
        \ENDFOR
        
        \IF{$\text{Random}(0,1) < \alpha$} 
            \FOR{each exchanged trait $T_i \in \mathcal{E}$}
                \FOR{each feature $p \in T_i$}
                    \STATE $R_p \leftarrow \max_{\mathbf{x} \in \mathcal{S}_c} \mathbf{x}^{(p)} - \min_{\mathbf{x} \in \mathcal{S}_c} \mathbf{x}^{(p)}$
                    \STATE $\hat{\mathbf{x}}^{(p)} \leftarrow \hat{\mathbf{x}}^{(p)} + 0.05 \cdot R_p \cdot \mathcal{N}(0,1)$
                \ENDFOR
            \ENDFOR
        \ENDIF
        
        \STATE $\mathcal{D}^+ \leftarrow \mathcal{D}^+ \cup \{(\hat{\mathbf{x}}, c)\}$
    \ENDFOR
\ENDFOR

\RETURN $\mathcal{D}^+$
\end{algorithmic}
\label{alg:sampling_algorithm}
\end{algorithm}

\section{Experimental Design}

In our experiments, we seek to answer the following questions:

\begin{itemize}
    \item \textbf{RQ1: }
    How does AxelSMOTE perform compared to existing sampling approaches for imbalanced classification tasks across different datasets and evaluation metrics?
    
    \item \textbf{RQ2: } 
    How sensitive is AxelSMOTE's performance to variations in its hyperparameters?
    
    \item \textbf{RQ3: }
    What is the individual contribution of each component within AxelSMOTE to the overall performance improvement?
    
    \item \textbf{RQ4: }
    How does the runtime performance of AxelSMOTE compare to baseline sampling approaches?
    
    \item \textbf{RQ5: }
    How do the synthetic data instances generated by AxelSMOTE compare visually and distributionally to those produced by baseline sampling methods?

\end{itemize}

\subsection{Datasets}

We evaluate AxelSMOTE performance on eight real-world datasets with diverse characteristics. These datasets are Wisconsin \citep{street1993nuclear}, Thyroids \citep{pang2019deep}, kc1 \citep{shirabad2005promise}, Ads \citep{das2018sparse}, ILPD \citep{ramana2012critical}, Glass \citep{dong2011new}, Page Blocks \citep{malerba1994page}, and Ecoli \citep{horton1996probabilistic}. The summary of these datasets is provided in Table \ref{tab:datasets}.

\begin{table}[htbp]
\centering
\caption{Dataset Summary}
\label{tab:datasets}
\begin{tabular}{l|c|c|c|l}
\hline
\textbf{Dataset} & \textbf{Type} & \textbf{Instances} & \textbf{Features} & \textbf{Description} \\
\hline
Wisconsin & Binary & 569 & 30 & Patient traits associated with breast cancer \\
\hline
Thyroid & Binary & 7,200 & 21 & Data related to thyroid disorders \\
\hline
Kc1 & Binary & 2,109 & 21 & Software defect prediction \\
\hline
Ads & Binary & 3,279 & 1,558 & Predicting possible internet \\ 
    &        &       &       & advertisements \\
\hline
ILPD & Binary & 583 & 10 & Data about the Indian liver patients \\
\hline
Glass & Multi-class & 214 & 9 & Identification of different glass types \\
\hline
Page Blocks & Multi-class & 5,473 & 10 & Classifying page layout blocks \\ 
           &             &       &    & detected by segmentation \\
\hline
Ecoli & Multi-class & 336 & 7 & Protein localization prediction \\
\hline
\end{tabular}
\end{table}

\subsection{Experimental Setups and Baselines}

We conduct a stratified training-testing split with multiple model executions, recording the average and standard deviation of the evaluation metrics. We further perform a data preprocessing step to normalize dataset features and fill in dummy values for missing features. As the classifier, we employ a Multi-Layer Perceptron (MLP) \citep{baum1988capabilities}, aligning with previous research \citep{karunasingha2023oc,kishanthan2025deep}. For baseline models, we employ state-of-the-art sampling approaches across three categories: oversampling, undersampling, and hybrid methods. 

\paragraph{Oversampling Methods:} Oversampling methods include approaches such as SMOTE \citep{chawla2002smote}, SVMSMOTE \citep{tang2008svms}, BorderlineSMOTE \citep{han2005borderline}, ADASYN \citep{he2008adasyn}, SMOTE-N \citep{chawla2002smote}, and SMOTENC \citep{chawla2002smote}. 

\paragraph{Undersampling Methods:} Undersampling approaches comprise Cluster Centroids \citep{lin2017clustering}, Condensed Nearest Neighbour \citep{hart1968condensed}, Edited Nearest Neighbour \citep{wilson2007asymptotic}, Instance Hardness Threshold \citep{smith2014instance}, Near Miss \citep{mani2003knn}, Neighbourhood Cleaning Rule \citep{laurikkala2001improving}, One Sided Selection \citep{kubat1997addressing}, and Tomek Links \citep{tomek1976two}. 

\paragraph{Hybrid Sampling Methods:} Hybrid sampling methods include SMOTEENN \citep{batista2004study} and SMOTETomek \citep{batista2003balancing}, which combine both oversampling and undersampling techniques.

\subsection{Evaluation Metrics}
We use F1 score\citep{luque2019impact} and balanced accuracy\citep{brodersen2010balanced} to evaluate the baselines and the model. The selection of these evaluation metrics is based on their higher sensitivity to the class imbalance problem. The equations of the evaluation metrics are as follows.

\begin{equation}
\text{F1-score} = 2 \times \frac{TP}{2TP + FP + FN}   
\end{equation}

\begin{equation}
\text{Balanced Accuracy} = \frac{1}{2} \left( \frac{TP}{TP + FN} + \frac{TN}{TN + FP} \right)
\end{equation}

Here TP, TN, FP, FN refer to true positives, true negatives, false positives, and false negatives, respectively.

\subsection{Model Hyper-parameters}

 Our experimental protocol follows a standard 80:20 data split strategy for training and testing phases. To ensure statistical robustness, we conduct 10 independent experimental runs for each model configuration using distinct random seeds. One training iteration consists of 200 epochs for all baseline methods and AxelSMOTE. However, for deep generative models (GANs and VAEs), we employ 10,000 epochs as these models require substantially longer training periods to converge effectively.

Our MLP classifier consists of one layer, with a hidden dimension of 64. We use a learning rate of 0.05 and a dropout rate of 0.0. Further, we employ the Adam algorithm \citep{kingma2014adam} as our optimiser. For each dataset, we optimize batch sizes by selecting from the range $\{500, 1000, 2500, 5000\}$, ensuring sufficient representation of minority class instances during the oversampling procedure. We search optimal hyperparameters for AxelSMOTE within the following ranges: \( k \in \{1, 2, 5, 6\} \), \( t \in \{1, 2, 4, 8, 12\} \), \( \theta \in \{0.2, 0.4, 0.6\} \), and \( \alpha \in \{0.2, 0.4, 0.6\} \). Further, we search for optimal hyperparameter configurations in our baselines and report their best results to ensure a fair comparison in our evaluations.

\subsection{System Resources and Implementation Details}

All the experiments were conducted in an environment with 32GB RAM and a 4-core CPU. We used the Python programming language \citep{van2007python} with  PyTorch \citep{imambi2021pytorch}, Scikit-learn \citep{kramer2016scikit}, Numpy \citep{oliphant2006guide}, and Pandas \citep{mckinney2011pandas} libraries for our implementations.

\begin{sidewaystable}[htbp]
\centering
\caption{F1-score ± standard deviation (\%) reported for imbalanced datasets. Best results are highlighted in \textbf{bold}.}
\label{tab:classification_f1-score}
\resizebox{\textwidth}{!}{%
\begin{tabular}{l|l|c|c|c|c|c|c|c|c|c}
\hline
\textbf{Category} & \textbf{Method} & Page-blocks & Glass & Wisconsin & Thyroid & Kc1 & Ads & ILPD & ecoli & Average \\
\hline
\multirow{6}{*}{Oversampling} 
& SMOTE & 64.53 ± 2.48 & 66.88 ± 8.34 & 97.83 ± 0.89 & 89.45 ± 2.88 & 59.47 ± 2.48 & 90.93 ± 1.57 & 64.04 ± 2.80 & \textbf{83.94 ± 5.22} & 77.13 \\
& SVMSMOTE & 65.20 ± 3.44 & 65.59 ± 8.34 & 97.29 ± 1.19 & 84.61 ± 2.09 & \textbf{65.11 ± 3.49} & 91.45 ± 1.58 & 62.57 ± 2.67 & 83.44 ± 3.95 & 76.91 \\
& BorderlineSMOTE & 62.58 ± 3.04 & 65.50 ± 10.88 & 96.83 ± 1.09 & 87.63 ± 2.64 & 60.38 ± 2.59 & 91.30 ± 1.70 & 63.74 ± 3.10 & 80.52 ± 5.13 & 76.06 \\
& ADASYN & 59.57 ± 6.92 & 63.40 ± 7.30 & 96.66 ± 1.50 & 88.79 ± 7.49 & 58.40 ± 2.64 & 90.41 ± 1.54 & \textbf{64.52 ± 3.30} & 81.32 ± 6.26 & 75.38 \\
& SMOTE-N & 65.30 ± 2.57 & 65.82 ± 10.92 & 96.95 ± 1.27 & 78.95 ± 3.11 & 64.30 ± 3.85 & 91.09 ± 1.55 & 62.67 ± 3.47 & 75.50 ± 6.41 & 75.07 \\
& SMOTENC & 65.29 ± 3.01 & 66.92 ± 9.26 & \textbf{97.93 ± 0.97} & 87.53 ± 5.73 & 59.80 ± 2.84 & 90.72 ± 1.35 & 63.23 ± 3.59 & 83.84 ± 4.85 & 76.91 \\
\hline
\multirow{8}{*}{Undersampling} 
& ClusterCentroids & 63.05 ± 3.12 & 57.70 ± 9.16 & 97.37 ± 1.30 & 78.80 ± 10.41 & 58.21 ± 11.28 & 86.14 ± 10.18 & 61.53 ± 3.10 & 79.68 ± 6.33 & 72.81 \\
& CondensedNearestNeighbour & 64.82 ± 3.29 & 57.65 ± 9.04 & 97.33 ± 1.35 & 80.04 ± 14.33 & 58.53 ± 11.19 & 86.68 ± 9.80 & 61.06 ± 4.64 & 78.06 ± 7.46 & 73.02 \\
& EditedNearestNeighbours & 59.66 ± 8.56 & 55.81 ± 9.81 & 97.41 ± 1.23 & 81.08 ± 11.82 & 58.80 ± 11.08 & 87.97 ± 6.55 & 61.96 ± 4.39 & 77.95 ± 7.40 & 72.58 \\
& InstanceHardnessThreshold & 56.67 ± 12.68 & 56.96 ± 9.05 & 96.60 ± 2.81 & 82.59 ± 2.10 & 59.05 ± 10.44 & 92.29 ± 1.19 & 59.22 ± 7.19 & 78.18 ± 7.38 & 72.70 \\
& NearMiss & 55.18 ± 12.66 & 55.94 ± 9.42 & 97.44 ± 1.22 & 78.64 ± 10.91 & 58.84 ± 10.89 & 85.09 ± 11.51 & 58.86 ± 7.55 & 77.80 ± 7.47 & 71.00 \\
& NeighbourhoodCleaningRule & 55.77 ± 12.36 & 56.46 ± 9.52 & 96.67 ± 2.78 & 83.52 ± 2.75 & 59.25 ± 10.33 & 92.27 ± 1.18 & 59.53 ± 7.13 & 78.28 ± 7.36 & 72.72 \\
& OneSidedSelection & 57.26 ± 12.09 & 56.30 ± 9.49 & 97.54 ± 1.17 & 78.84 ± 10.71 & 58.97 ± 10.73 & 85.67 ± 11.24 & 58.97 ± 7.59 & 77.99 ± 7.45 & 71.44 \\
& TomekLinks & 59.07 ± 12.02 & 56.61 ± 9.54 & 97.60 ± 1.13 & 79.03 ± 10.52 & 59.09 ± 10.57 & 92.54 ± 1.37 & 59.04 ± 7.37 & 78.16 ± 7.43 & 72.64 \\
\hline
\multirow{2}{*}{Hybrid} 
& SMOTEENN & 62.37 ± 2.65 & 61.81 ± 7.71 & 97.93 ± 1.15 & 83.55 ± 6.13 & 59.63 ± 2.02 & 91.43 ± 1.02 & 63.34 ± 2.35 & 83.78 ± 4.97 & 75.48 \\
& SMOTETomek & 62.61 ± 2.75 & 62.01 ± 8.07 & 97.79 ± 1.07 & 84.15 ± 8.09 & 59.77 ± 2.84 & 91.04 ± 1.29 & 62.13 ± 3.35 & 83.84 ± 5.20 & 75.42 \\
\hline
\multirow{1}{*}{Proposed} 
& AxelSMOTE & \textbf{77.88 ± 3.21} & \textbf{69.12 ± 8.86} & 97.56 ± 0.89 & \textbf{89.63 ± 5.36} & 62.74 ± 2.69 & \textbf{92.68 ± 1.08} & 64.13 ± 3.34 & 82.24 ± 5.00 & \textbf{79.50} \\
\hline
\end{tabular}
}
\end{sidewaystable}

\begin{sidewaystable}[htbp]
\centering
\caption{Balanced Accuracy ± standard deviation (\%) reported for imbalanced datasets. Best results are highlighted in \textbf{bold}.}
\label{tab:classification_balanced_accuracy}
\resizebox{\textwidth}{!}{%
\begin{tabular}{l|l|c|c|c|c|c|c|c|c|c}
\hline
\textbf{Category} & \textbf{Method} & Page-blocks & Glass & Wisconsin & Thyroid & Kc1 & Ads & ILPD & ecoli & Average \\
\hline
\multirow{6}{*}{Oversampling} 
& SMOTE & 91.53 ± 1.69 & 69.64 ± 8.08 & 97.86 ± 0.88 & 94.29 ± 3.36 & 71.54 ± 4.07 & 92.87 ± 1.43 & 70.07 ± 3.15 & 85.19 ± 5.55 & 84.12 \\
& SVMSMOTE & 90.96 ± 2.17 & 66.30 ± 6.99 & 97.59 ± 1.01 & 88.06 ± 6.17 & 70.32 ± 3.46 & 93.04 ± 1.06 & 67.28 ± 2.89 & 83.72 ± 4.07 & 82.16 \\
& BorderlineSMOTE & 91.40 ± 1.83 & 66.67 ± 10.67 & 97.24 ± 0.98 & 93.38 ± 4.18 & 71.47 ± 3.67 & 93.19 ± 1.37 & 70.26 ± 3.31 & 81.82 ± 4.99 & 83.18 \\
& ADASYN & 88.81 ± 6.16 & 64.53 ± 6.47 & 97.03 ± 1.15 & 95.55 ± 2.10 & 72.09 ± 5.74 & 93.08 ± 1.21 & \textbf{70.86 ± 3.39} & 81.64 ± 6.47 & 82.95 \\
& SMOTE-N & 88.15 ± 3.46 & 66.03 ± 10.64 & 96.94 ± 1.28 & 86.16 ± 3.08 & 63.85 ± 3.74 & 93.14 ± 1.08 & 63.78 ± 3.87 & 85.17 ± 5.85 & 80.40 \\
& SMOTENC & 91.75 ± 2.09 & 68.05 ± 9.72 & \textbf{97.92 ± 0.97} & 94.87 ± 2.99 & 71.00 ± 3.77 & 93.10 ± 1.20 & 68.84 ± 4.29 & \textbf{85.54 ± 5.41} & 83.88 \\
\hline
\multirow{8}{*}{Undersampling} 
& ClusterCentroids & 87.77 ± 2.40 & 61.06 ± 8.43 & 97.51 ± 1.14 & 88.01 ± 3.61 & 63.75 ± 8.01 & 90.16 ± 4.68 & 66.80 ± 3.25 & 82.14 ± 6.08 & 79.65 \\
& CondensedNearestNeighbour & 86.10 ± 3.47 & 61.65 ± 7.71 & 97.48 ± 1.15 & 92.00 ± 5.21 & 63.82 ± 7.89 & 90.35 ± 4.50 & 64.43 ± 5.32 & 79.52 ± 6.78 & 79.42 \\
& EditedNearestNeighbours & 72.90 ± 19.70 & 59.19 ± 9.64 & 97.55 ± 1.07 & 87.42 ± 8.17 & 64.93 ± 7.85 & 91.02 ± 2.70 & 65.65 ± 5.27 & 79.40 ± 6.73 & 77.26 \\
& InstanceHardnessThreshold & 65.03 ± 18.01 & 60.65 ± 9.27 & 96.93 ± 2.19 & 85.80 ± 2.86 & 64.28 ± 7.69 & 93.09 ± 1.62 & 62.96 ± 7.77 & 75.86 ± 7.30 & 75.58 \\
& NearMiss & 73.52 ± 17.08 & 59.60 ± 9.41 & 97.55 ± 1.07 & 86.77 ± 7.25 & 64.31 ± 7.85 & 89.58 ± 5.31 & 62.84 ± 8.08 & 79.27 ± 6.82 & 76.68 \\
& NeighbourhoodCleaningRule & 62.41 ± 18.32 & 59.87 ± 9.38 & 96.99 ± 2.18 & 83.33 ± 4.81 & 64.46 ± 7.66 & 92.89 ± 1.57 & 63.26 ± 7.76 & 79.63 ± 6.71 & 75.35 \\
& OneSidedSelection & 71.29 ± 16.04 & 59.70 ± 9.60 & 97.63 ± 1.03 & 84.77 ± 7.92 & 64.16 ± 7.76 & 89.80 ± 5.17 & 62.78 ± 7.86 & 79.38 ± 6.79 & 76.19 \\
& TomekLinks & 69.48 ± 15.35 & 59.91 ± 8.89 & 97.68 ± 1.00 & 83.63 ± 7.86 & 64.02 ± 7.67 & 92.41 ± 1.81 & 62.70 ± 7.86 & 79.48 ± 6.76 & 76.16 \\
\hline
\multirow{2}{*}{Hybrid} 
& SMOTEENN & 91.21 ± 1.74 & 66.75 ± 9.01 & 97.92 ± 1.04 & 92.79 ± 3.43 & \textbf{72.25 ± 2.96} & \textbf{93.52 ± 0.95} & 69.76 ± 2.86 & 85.08 ± 5.24 & 83.66 \\
& SMOTETomek & 91.24 ± 1.76 & 66.23 ± 9.22 & 97.82 ± 1.03 & 92.95 ± 3.49 & 71.85 ± 3.02 & 93.19 ± 1.18 & 67.03 ± 4.62 & 83.09 ± 5.46 & 82.93 \\
\hline
\multirow{1}{*}{Proposed} 
& AxelSMOTE & \textbf{92.03 ± 2.72} & \textbf{70.16 ± 9.38} & 97.70 ± 0.68 & \textbf{97.69 ± 1.14} & 68.47 ± 3.58 & 93.45 ± 1.26 & 70.15 ± 4.34 & 84.07 ± 4.58 & \textbf{84.22} \\
\hline
\end{tabular}
}
\end{sidewaystable}

\section{Experiments}

\subsection{Imbalanced Classification Performance (RQ1)}

We compare the imbalanced classification performance of AxelSMOTE with state-of-the-art sampling approaches. The F1-score and balanced accuracy metric comparisons are provided in Table \ref{tab:classification_f1-score}, and Table \ref{tab:classification_balanced_accuracy}, respectively. AxelSMOTE demonstrates superiority in both F1-score and balanced accuracy, achieving the highest average performance among all methods. AxelSMOTE consistently outperforms traditional SMOTE-based methods across both metrics. When compared to the original SMOTE method, AxelSMOTE shows improvements of 2.37\% in F1-score on average. Against more sophisticated SMOTE variants like BorderlineSMOTE and SVMSMOTE, AxelSMOTE also maintains consistent advantages. Further, AxelSMOTE significantly outperforms all undersampling and hybrid sampling techniques. Additionally, the reported standard deviations indicate that AxelSMOTE maintains relatively stable performance across different experimental runs, suggesting robust and reliable performance.

\subsection{Hyperparameter Sensitivity Analysis (RQ2)}

\begin{figure}[h!]
    \centering
    \includegraphics[width=\textwidth]{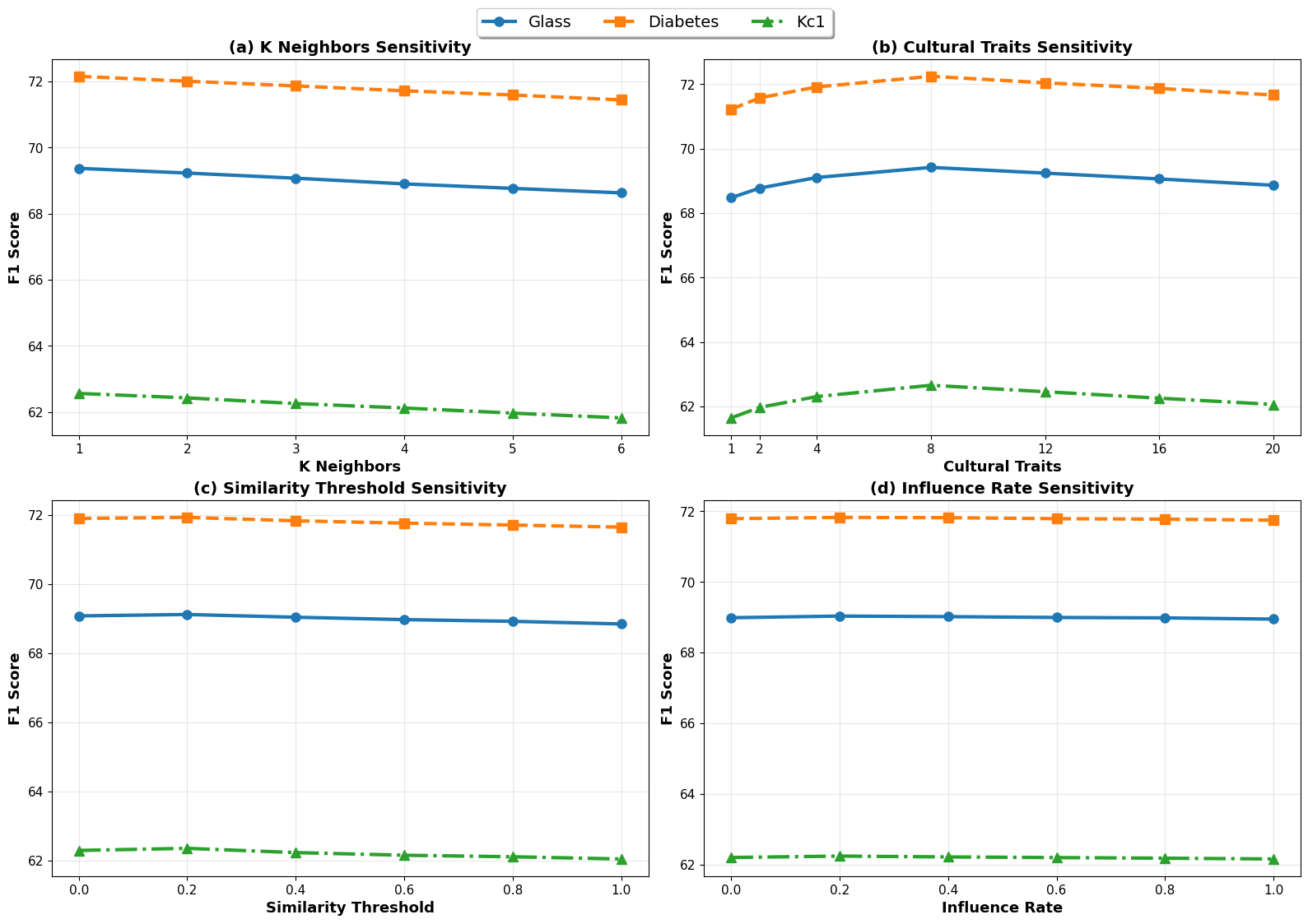}  
    \caption{Hyperparameter Sensitivity Analysis}
    \label{fig:hyperparameter_sensitivity}
\end{figure}

We conduct a parameter sensitivity analysis to evaluate four key hyperparameters in AxelSMOTE. The classification performance under parameter ranges is depicted in Figure \ref{fig:hyperparameter_sensitivity}. K-neighbours exhibits the highest sensitivity, with performance consistently decreasing as $k$ increases, with an optimal range being $[1, 2]$ across all datasets. The number of functional traits shows dataset-dependent sensitivity. The Glass dataset peaks at 1-4 traits, the KC1 benefits from 4-12 traits, while Diabetes remains stable across 1-20 traits. This suggests cultural complexity should match dataset characteristics. Similarity threshold and influence rate have optimal ranges of 0.2-0.4.

\subsection{Component Analysis (RQ3)}

We provide an ablation study to demonstrate the individual contribution of key components in the AxelSMOTE algorithm by systematically removing each component and measuring performance degradation across three datasets. The results are depicted in Table \ref{tab:ablation_study}. 

\begin{table}[h]
\centering
\begin{tabular}{l|c|c|c}
\hline
\textbf{Method} & \textbf{Wisconsin} & \textbf{ADs} & \textbf{ILPD} \\
\hline
W/o feature trait grouping & $97.36 \pm 1.14$ & $92.25 \pm 1.08$ & $62.61 \pm 2.86$ \\
\hline
W/o similarity filtering & $97.18 \pm 1.23$ & $91.43 \pm 1.03$ & $61.48 \pm 4.11$ \\
\hline
W/o beta distribution blending & $97.17 \pm 0.75$ & $91.17 \pm 0.68$ & $60.72 \pm 4.69$ \\
\hline
W/o diversity injection & $97.24 \pm 1.41$ & $92.46 \pm 1.05$ & $62.26 \pm 5.31$ \\
\hline
AxelSMOTE & $\boldsymbol{97.56 \pm 0.89}$ & $\boldsymbol{92.68 \pm 1.08}$ & $\boldsymbol{64.13 \pm 3.34}$ \\
\hline
\end{tabular}
\caption{Ablation study for different components in AxelSMOTE. Classification F1-score  ± standard deviation is reported. Best results are highlighted in \textbf{bold}.}
\label{tab:ablation_study}
\end{table}

The full AxelSMOTE algorithm achieves best performance through the synergistic interaction of all components, with each component contributing to different aspects of synthetic sample generation. The beta distribution blending component has the highest performance impact, while other components show moderate impacts. We believe this is because beta distribution blending enhances the core mathematical interpolation process that generates synthetic samples, replacing the simple interpolation used in traditional sampling models.

\subsection{Runtime Efficiency Comparison (RQ4)}

\begin{figure}[htbp]
    \centering
    \includegraphics[width=\textwidth]{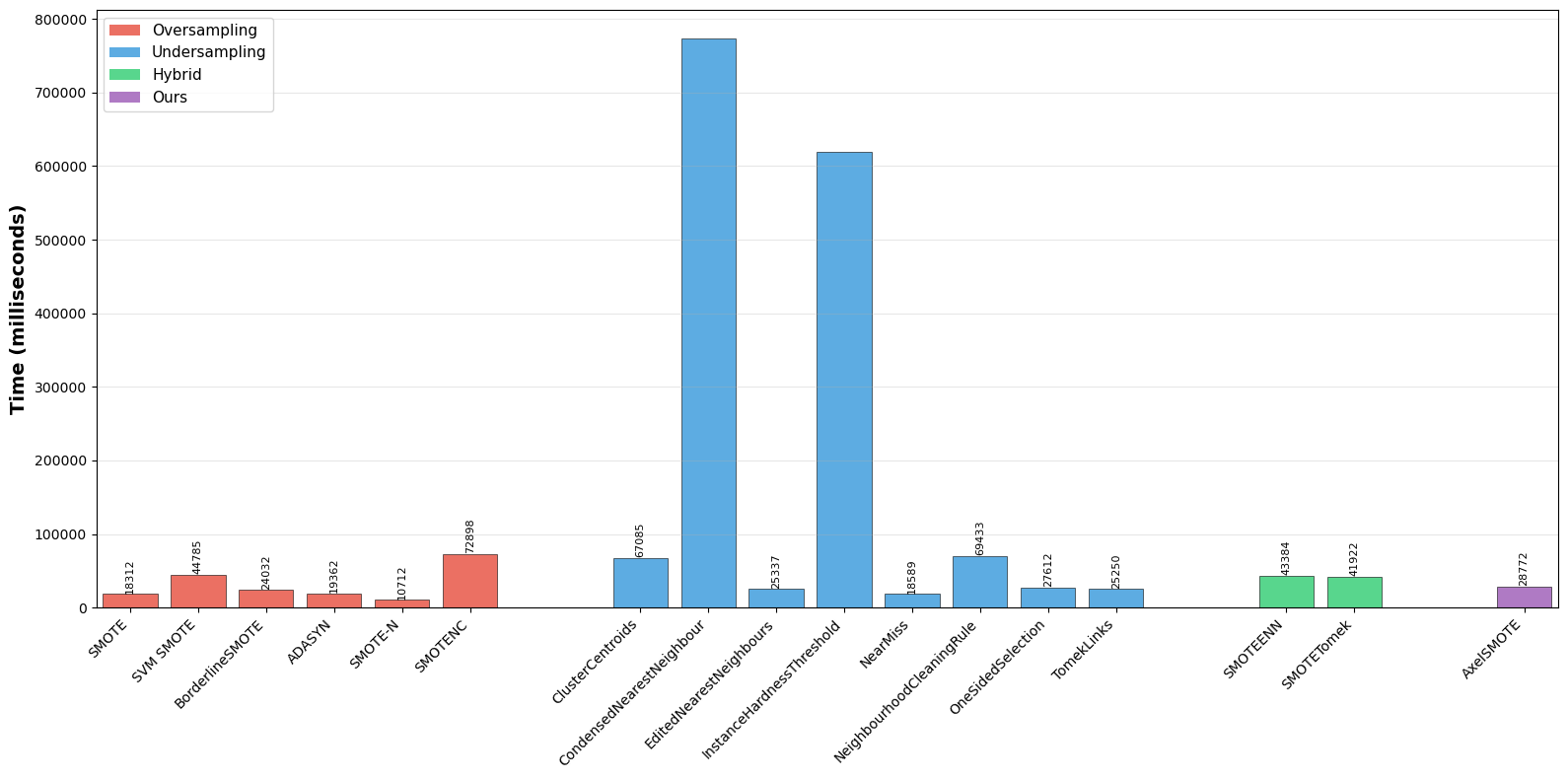} 
    \caption{Runtime Analysis (per training cycle) for Thyroid Dataset}
    \label{fig:runtime_analysis}
\end{figure}

Figure \ref{fig:runtime_analysis} shows the runtime comparison of AxelSMOTE with other sampling approaches. AxelSMOTE demonstrates competitive runtime efficiency, including hybrid and undersampling techniques. While slightly slower than basic SMOTE variants, it outperforms computationally expensive oversampling methods like SMOTENC. This positions AxelSMOTE as a balanced solution, offering improved synthetic sample generation without excessive computational overhead, making it practical for real-world data applications.

\subsection{Visualisation: Synthetic Data Quality (RQ5)}

\begin{figure}[htbp]
    \centering
    \begin{tabular}{ccc}
        \subcaptionbox{SMOTE\label{fig:img1}}{%
            \includegraphics[width=0.3\textwidth]{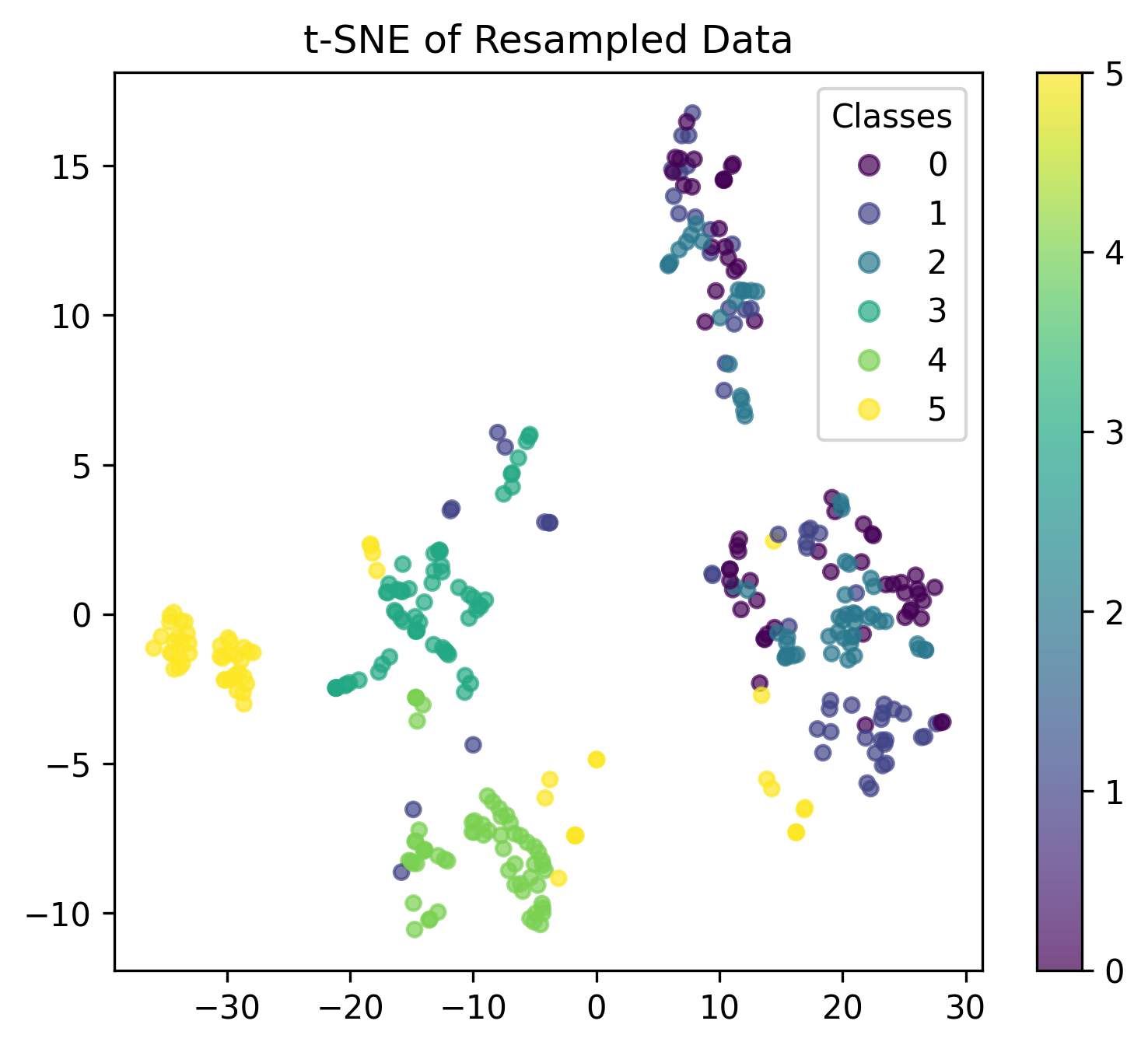}} &
        \subcaptionbox{SVMSMOTE\label{fig:img2}}{%
            \includegraphics[width=0.3\textwidth]{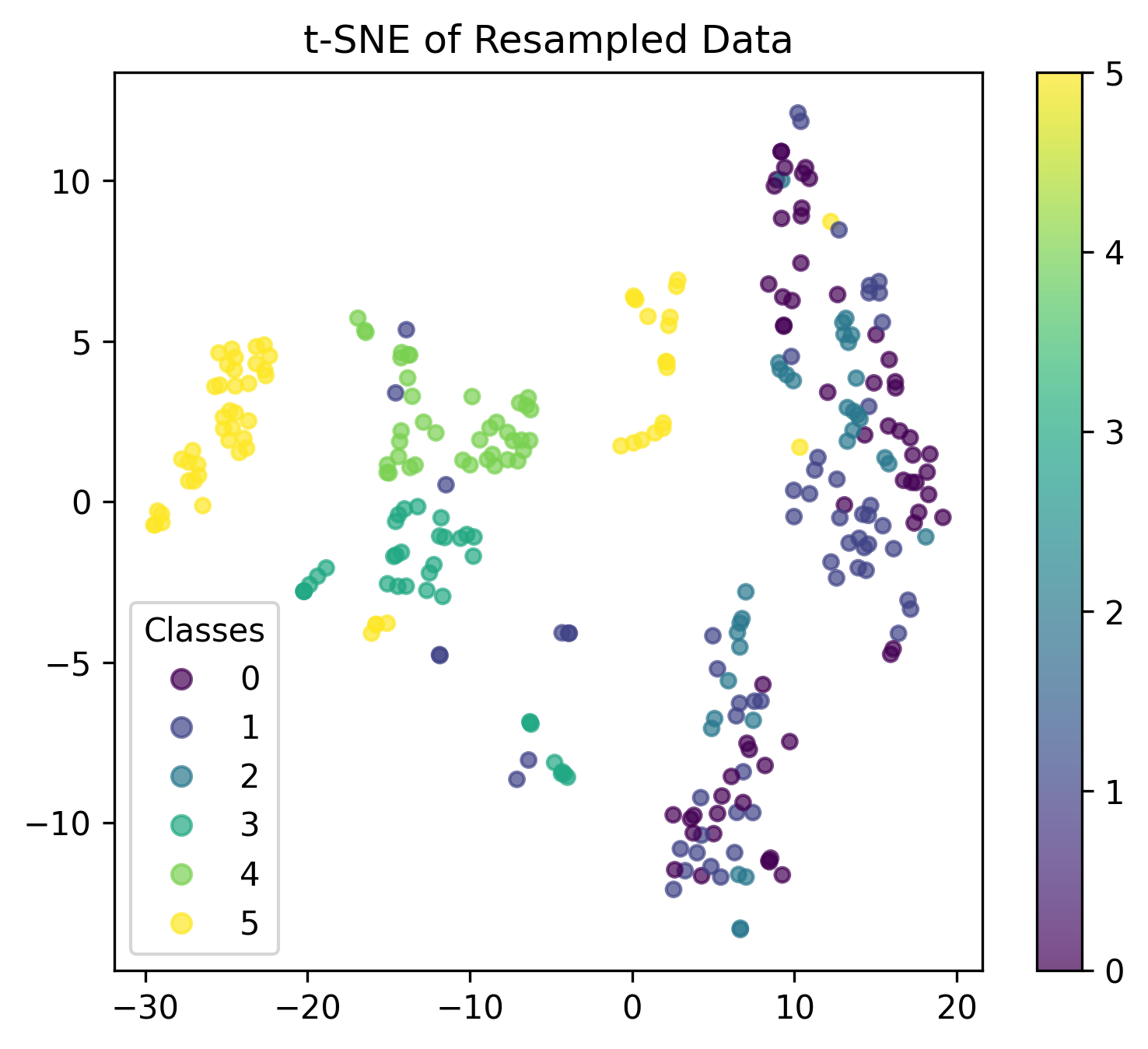}} &
        \subcaptionbox{BorderlineSMOTE\label{fig:img3}}{%
            \includegraphics[width=0.3\textwidth]{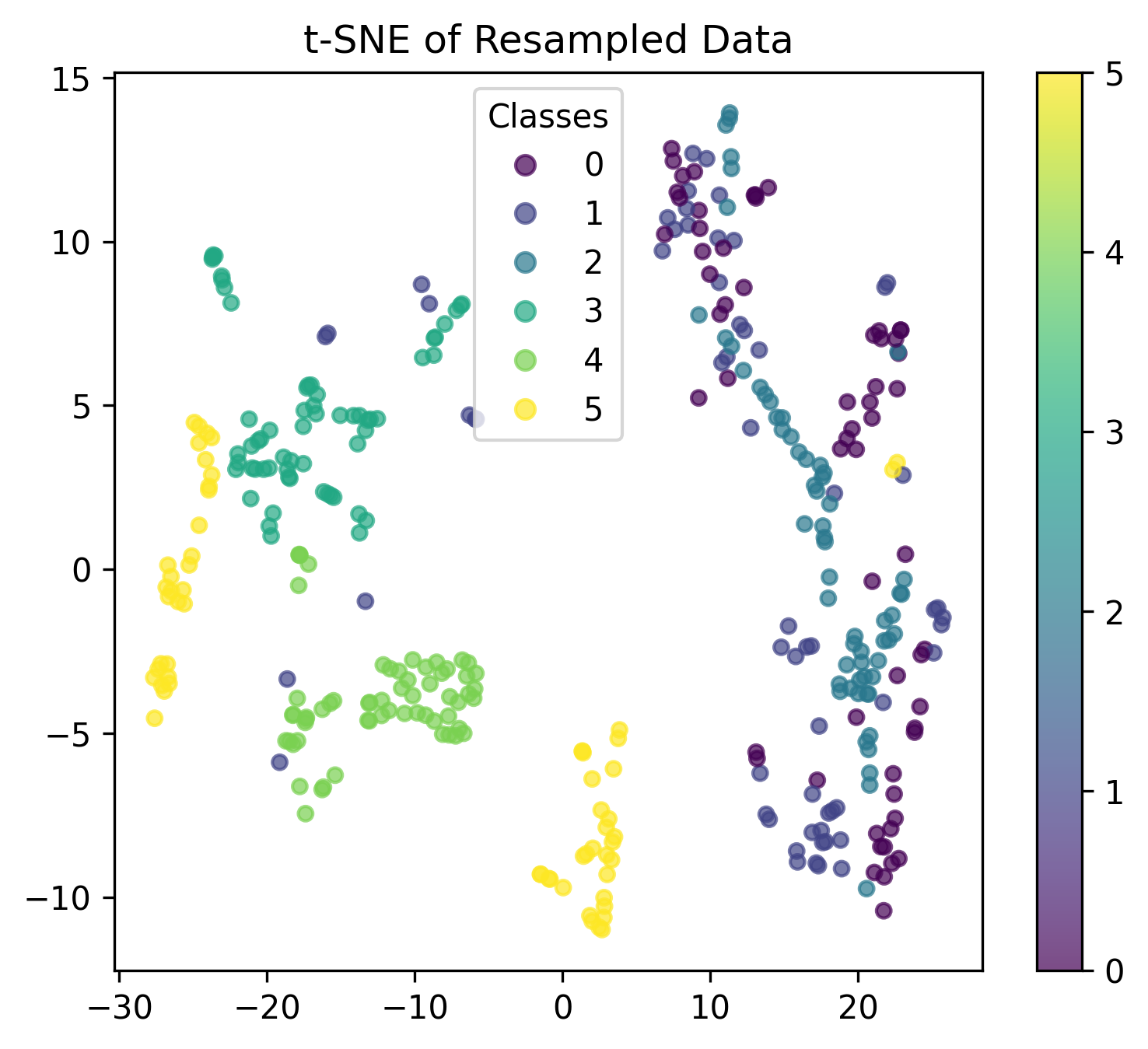}} \\
        
        \subcaptionbox{Condensed Nearest Neighbour\label{fig:img4}}{%
            \includegraphics[width=0.3\textwidth]{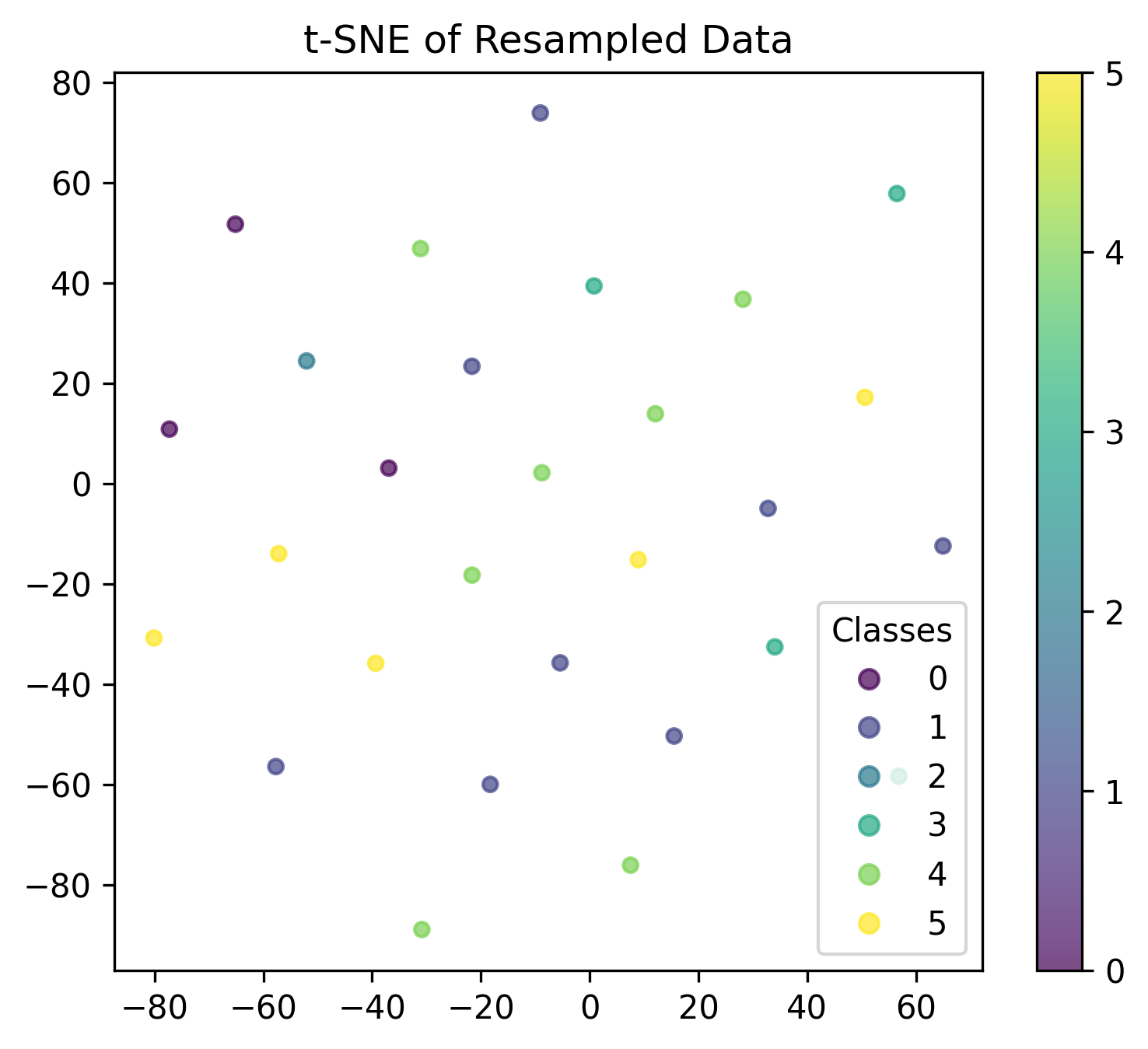}} &
        \subcaptionbox{Neighbourhood Cleaning Rule\label{fig:img5}}{%
            \includegraphics[width=0.3\textwidth]{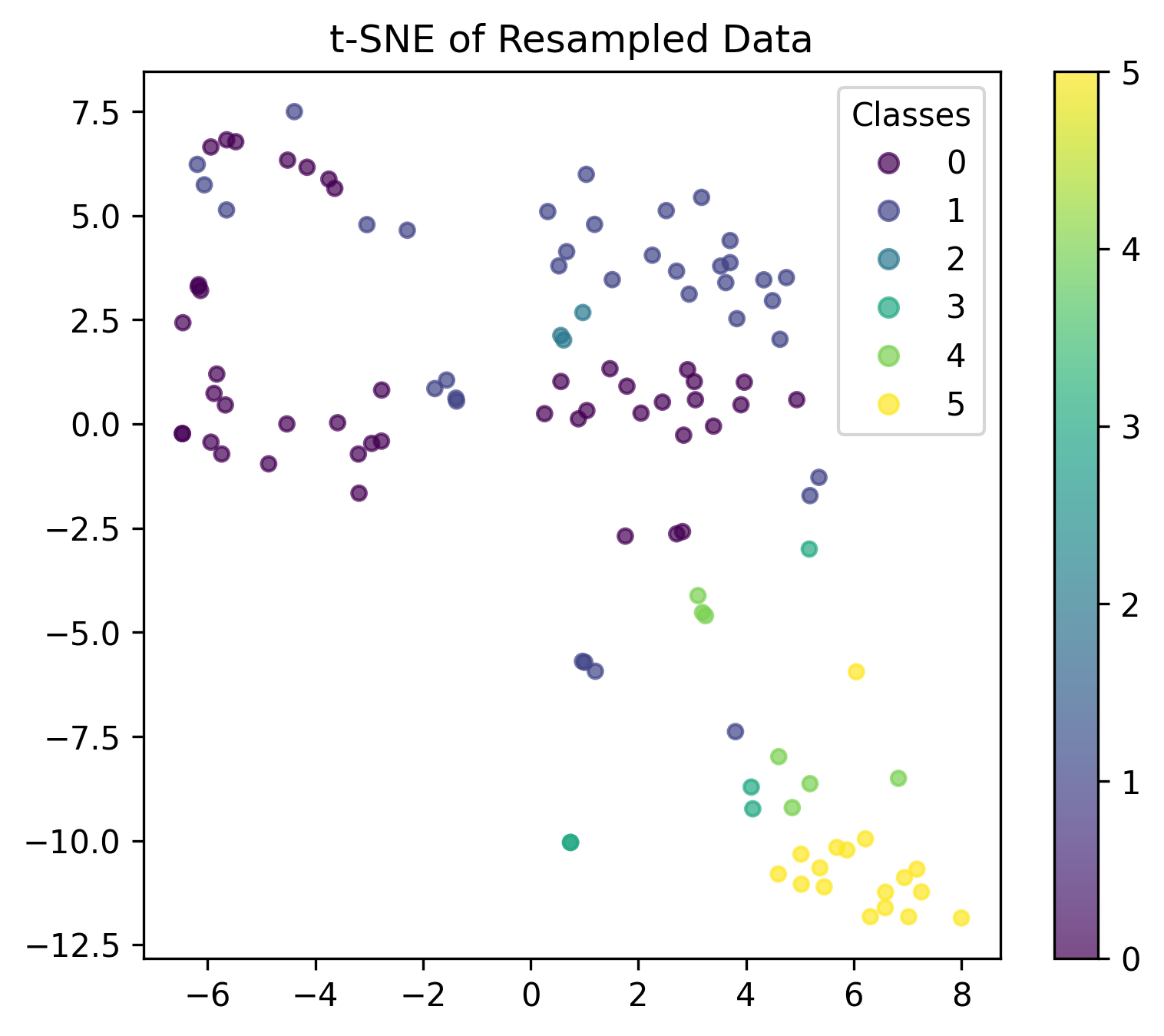}} &
        \subcaptionbox{One-Sided Selection\label{fig:img6}}{%
            \includegraphics[width=0.3\textwidth]{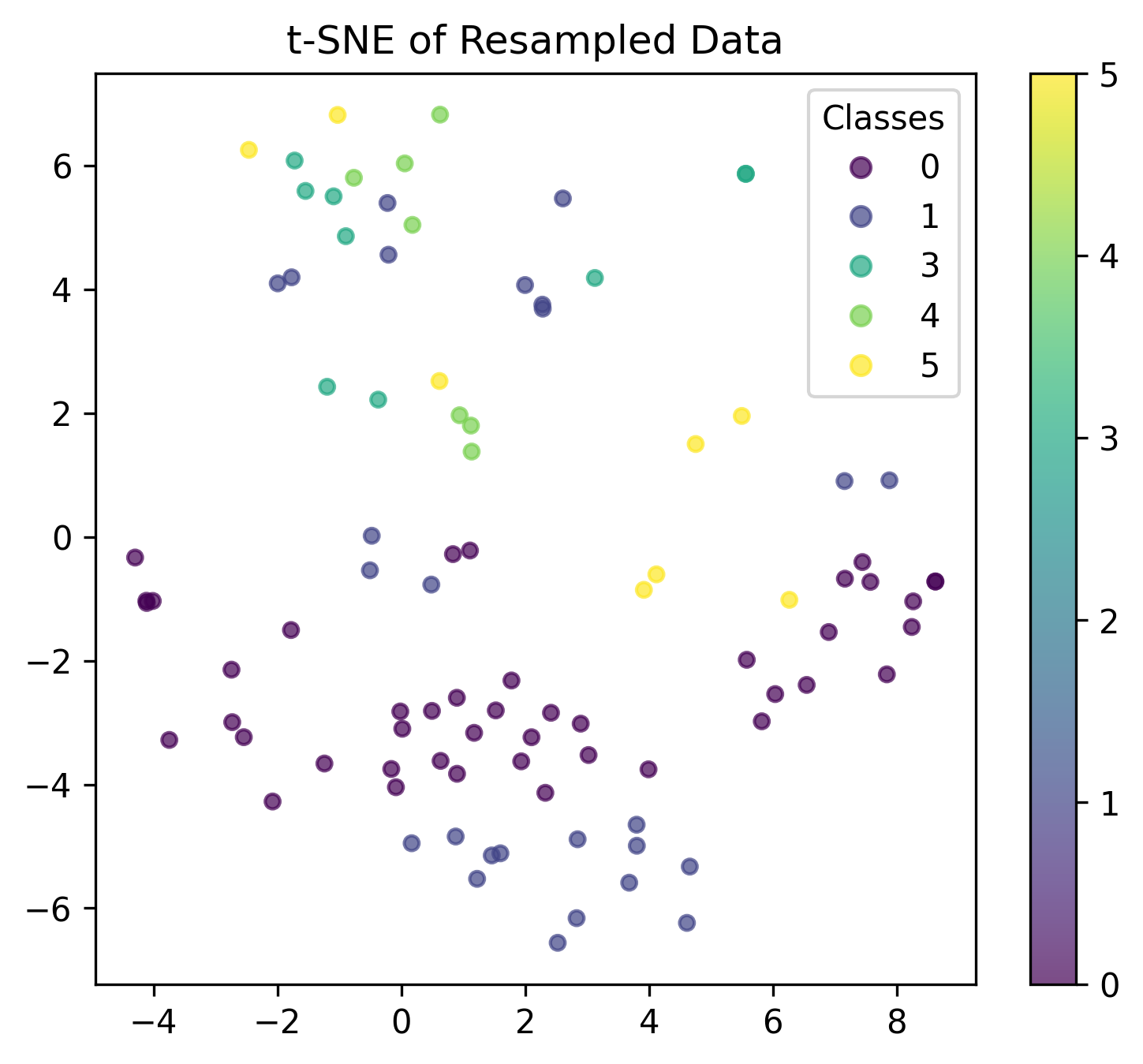}} \\
        
        \subcaptionbox{SMOTEENN\label{fig:img7}}{%
            \includegraphics[width=0.3\textwidth]{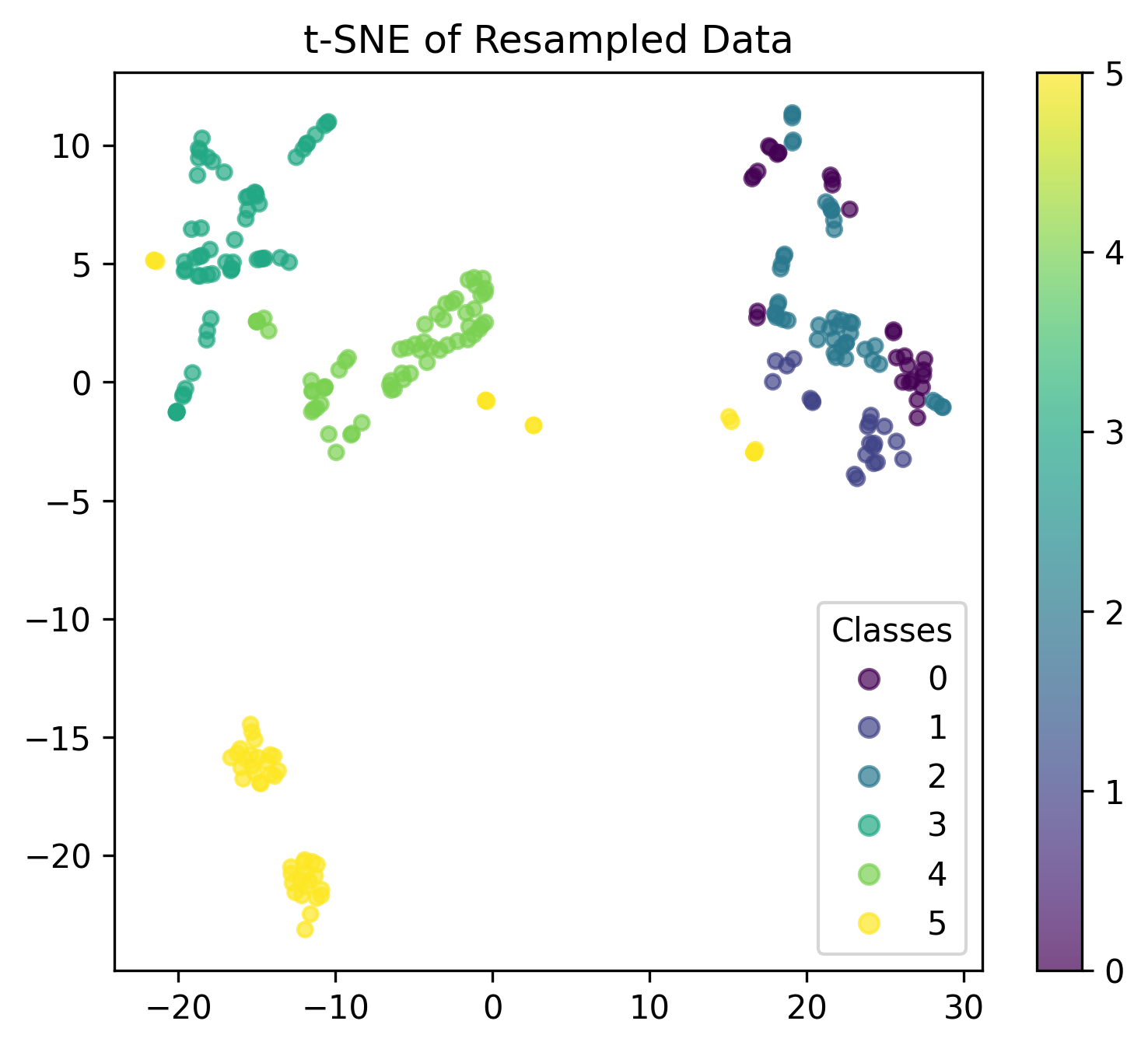}} &
        \subcaptionbox{SMOTETomek\label{fig:img8}}{%
            \includegraphics[width=0.3\textwidth]{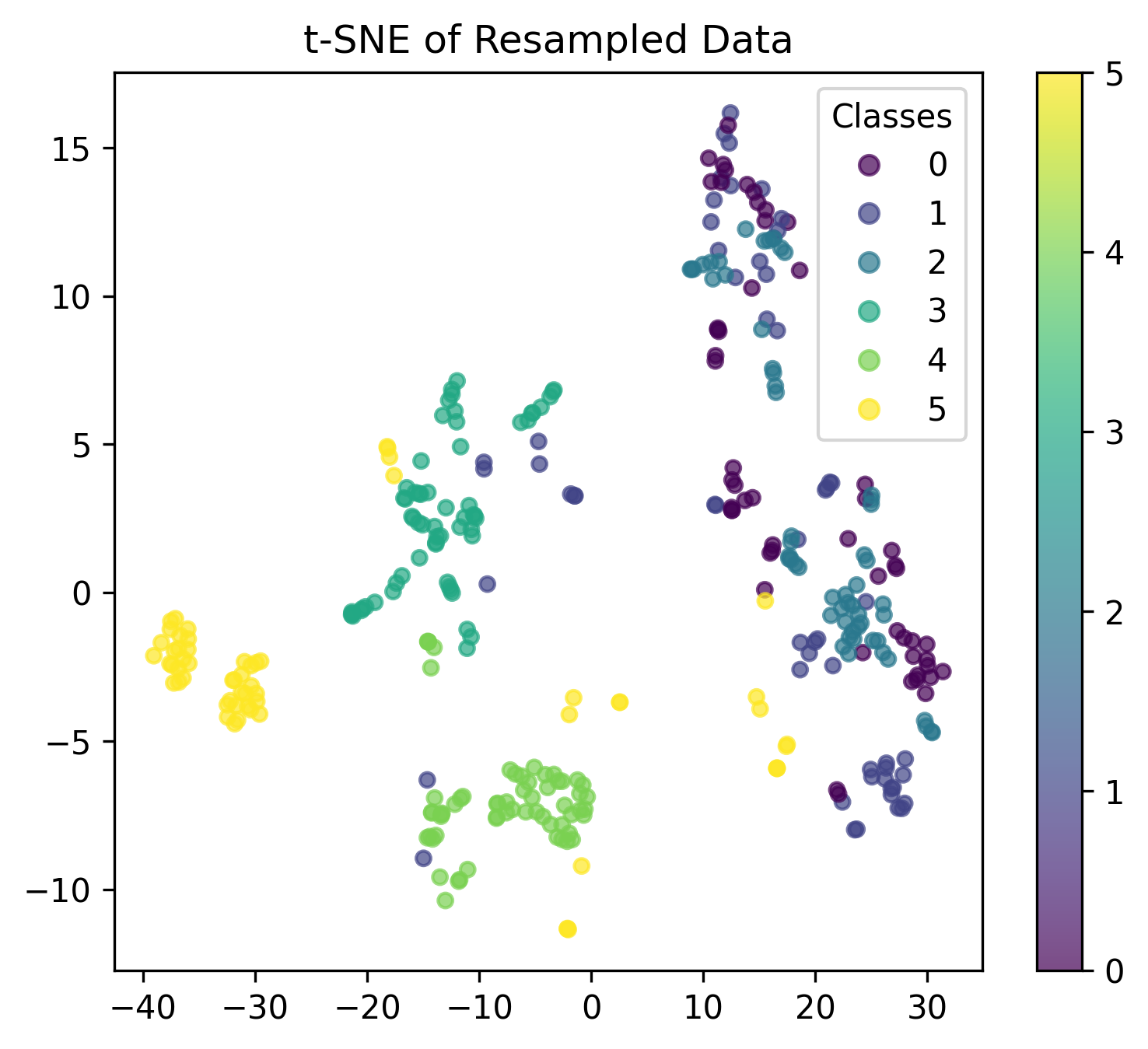}} &
        \subcaptionbox{AxelSMOTE\label{fig:img9}}{%
            \includegraphics[width=0.3\textwidth]{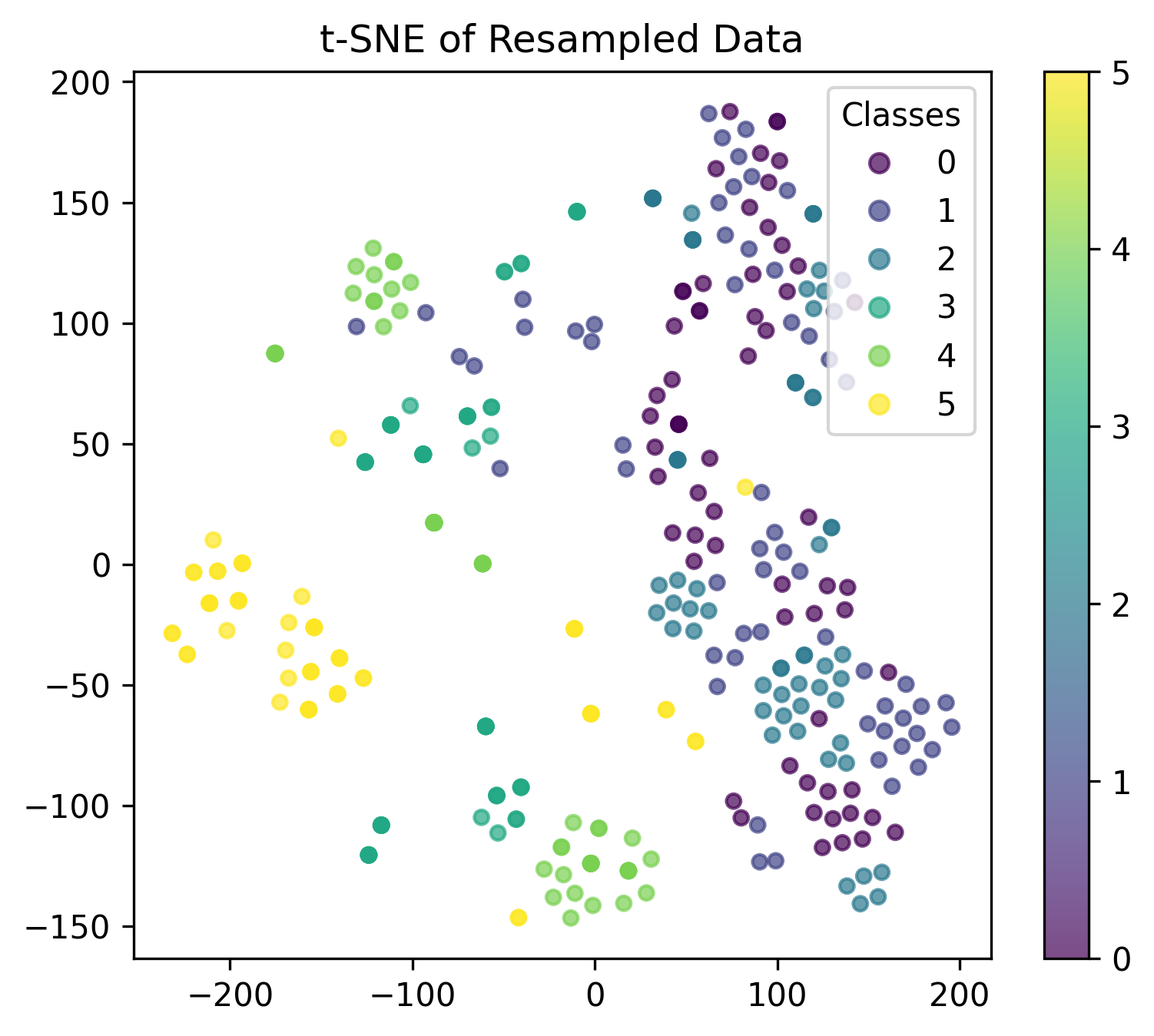}} \\
    \end{tabular}
    \caption{t-SNE visualizations of the \textit{glass} dataset after applying various resampling techniques: (a) SMOTE, (b) SVMSMOTE, (c) BorderlineSMOTE, (d) Condensed Nearest Neighbour, (e) Neighbourhood Cleaning Rule, (f) One-Sided Selection, (g) SMOTEENN, (h) SMOTETomek, and (i) AxelSMOTE.}
    \label{fig:oversampling_quality}
\end{figure}

To assess the quality of synthetic samples generated by different resampling techniques, we conducted t-SNE visualizations \citep{maaten2008visualizing} on the Glass dataset, as shown in Figure \ref{fig:oversampling_quality}. Traditional SMOTE variants show varying separation quality, with SMOTE exhibiting notable inter-class overlap, SVMSMOTE providing slightly better boundaries, and BorderlineSMOTE focusing on boundary regions but potentially introducing noise. Undersampling approaches create sparser distributions with potential information loss, while hybrid methods SMOTEENN and SMOTETomek achieve moderate separation with reduced overlap through noise reduction techniques.

 Most notably, AxelSMOTE demonstrates superior visual characteristics across multiple dimensions. The method produces the most distinct class separation with minimal inter-class overlap, while maintaining cohesive intra-class clustering. The synthetic samples appear to follow natural data distributions, suggesting that the agent-based cultural exchange mechanism effectively preserves feature correlations and semantic relationships. The clear cluster boundaries and tight groupings indicate that AxelSMOTE generates realistic synthetic instances that enhance class discriminability without introducing artificial noise.

\section{Conclusion, Limitations, and Future Work}

In this work, we proposed a novel agent-based oversampling algorithm for class imbalance. Our algorithm is rooted in Axelrod's cultural dissemination model, which simulates the dynamics of social influence by allowing agents to interact and adopt each other’s attributes based on their similarity. By adapting this mechanism to the data domain, minority class samples are treated as agents that interact within a feature space, generating synthetic instances that preserve intrinsic data characteristics while enhancing diversity. Extensive experiments validate the superiority of the proposed method over state-of-the-art sampling approaches. 

Currently, our algorithm requires tuning of four hyperparameters. Our sensitivity analysis has identified optimal ranges for these parameters, which will reduce the need for manual tuning when applying our method to new datasets. However, in our future work, we plan to develop a data-driven approach to learn these parameters. Further, we plan to extend our work to other data types, such as time series data and images.

\section*{Data Availability Statement}

All data supporting the findings of this study are publicly available and have been properly cited within the article.

\section*{Conflict of Interest}

The authors declare that they have no conflict of interest.

\bibliography{references}

\end{document}